\documentclass[nohyperref]{article}

\usepackage{microtype}
\usepackage{graphicx}
\usepackage{booktabs} 

\usepackage{hyperref}

 \usepackage[accepted]{icml2022}

\usepackage{amsmath}
\usepackage{amssymb}
\usepackage{mathtools}
\usepackage{amsthm}
\usepackage{subfig}

\usepackage[capitalize,noabbrev]{cleveref}

\theoremstyle{plain}

\theoremstyle{definition}

\theoremstyle{remark}

\usepackage[textsize=tiny]{todonotes}

\icmltitlerunning{Easy Batch Normalization}

\begin{document}

\twocolumn[
\icmltitle{Easy Batch Normalization}



\icmlsetsymbol{equal}{*}

\begin{icmlauthorlist}
\icmlauthor{Arip Asadulaev}{Itmo,Airi}
\icmlauthor{Alexander Panfilov}{Tubingen}
\icmlauthor{Andrey Filchenkov}{Itmo}
\end{icmlauthorlist}

\icmlaffiliation{Itmo}{ITMO University, Saint-Petersburg, Russia}
\icmlaffiliation{Airi}{Artificial Intelligence Research Institute, Moscow, Russia}
\icmlaffiliation{Tubingen}{Tubingen Universitat, Germany}

\icmlcorrespondingauthor{Arip Asadulaev}{aripasadulaev@itmo.ru}

\icmlkeywords{Machine Learning, ICML}

\vskip 0.3in



\icmlkeywords{Machine Learning, ICML}

\vskip 0.3in
]



\printAffiliationsAndNotice{} 

\begin{abstract}
It was shown that adversarial examples improve object recognition. But what about their opposite side, easy examples? Easy examples are samples that the machine learning model classifies correctly with high confidence. In our paper, we are making the first step toward exploring the potential benefits of using easy examples in the training procedure of neural networks. We propose to use an auxiliary batch normalization for easy examples for the standard and robust accuracy improvement.

\end{abstract}\vspace{-4mm}
\vspace{-5mm}\section{Introduction}
Samples from the training dataset that the neural network \textbf{classifies correctly} with high confidence after the first training epoch are called easy examples~\cite{deepmemorization}. Recent results showed that easy examples could be removed from the training dataset without influencing the model generalization~\cite{EmpiricalEasy}. Although, we suppose that we can utilize easy examples to improve neural classifiers' standard and robust accuracy by enhancing the training procedure. 

However, the number of easy examples depends on the chosen classifier and dataset. To avoid this limitation and obtain more easy examples, we propose to use \textbf{targeted adversarial attacks}~\cite{YuanHZL19} with a ground truth target label to turn all training samples into their easy copies. In Section~(\ref{sec-study}), we show that targeted adversarial attacks actually make samples easy to classify. When we obtain easy examples, \textbf{we propose to use an auxiliary batch norm} to address them, similarly as it was proposed for adversarial examples~\cite{advprop}. Our hypothesis is that network's \textbf{layers trained with easy examples can "clean up" the adversarial data points} that lie away from the original data manifold and help to achieve a practical trade-off between the standard and robust accuracy. We provide experiments and show that easy batch normalization achieves improvement in robust accuracy similar to the original AdvProp~\cite{advprop} and even outperforms it in the standard accuracy on FashionMNIST and CIFAR-10~\cite{Krizhevsky09learningmultiple} datasets.

\section{Background}

\textbf{Adversarial examples:} Adversarial attacks are intended to ``fool`` the classifier~\cite{DBLP:journals/corr/SzegedyZSBEGF13, DBLP:conf/ccs/PapernotMGJCS17, YuanHZL19, DBLP:conf/iclr/SchottRBB19}. Using the sample $x$, the target label $y$, and model with parameters $\theta$, we can apply Projected Gradient Descent (PGD)~\cite{MadryMSTV18} iterations to get the adversarial example: 
\vspace{-2mm}
\begin{equation}
    {x}_{i+1}=\operatorname{Proj}_{(x, \varepsilon)}\left[{x}_{i}+\alpha \operatorname{sign}\left(-\nabla_{x} L\left(\theta, {x}_{i}, y_{i}\right)\right)\right]
    \label{eq:easyattack}
\end{equation}\vspace{-3mm}

Where, $\operatorname{Proj}_{(x, \varepsilon)}$  is a projection operator onto the $l_{\inf}$  ball of radius $\varepsilon$ around the original image $x_i$. 
The underlying data distribution of the adversarial examples is different from the real training data, and training on these examples improves the robust accuracy but decreases the standard one. However, recent results show applications of the adversarial examples for model improvement. The~\cite{advprop} proposed to use the adversarial examples in the separate batch normalization (i.e., AdvProp) and showed that it could improve the standard and adversarial accuracy of a model.

\textbf{Easy examples:} There is no formal definition for easy examples yet. In our study, we follow the simplest notation proposed in the \citep{deepmemorization}: instances that the model classifies accurately with high confidence are called easy examples.

\section{Easy batch normalization}
\label{sec-study}
The number of easy examples depends on the chosen classifier and dataset. We propose to obtain easy examples by gradient perturbations to avoid this limitation, such examples we called perturbed easy examples. Using gradient descent, we can add \textit{``easy features``} to the data similarly to how we add adversarial features. The only difference is that the $y_{i}$ in (\ref{eq:easyattack}) must be equal to the ground truth class label. Such examples are similar to the gradient-based preferable inputs that were used to understand and visualize neural network features~\cite{SimonyanVZ13, YosinskiCNFL15, MahendranV16}.

\textbf{Do gradients provide easy examples?} With this experiment, we justify that gradient perturbations are indeed adding to the image features that are easy to classify correctly. 

\vspace{-2mm}\begin{figure}[!htb]
\centering
\minipage{0.22\textwidth}
  \includegraphics[width=0.9\linewidth]{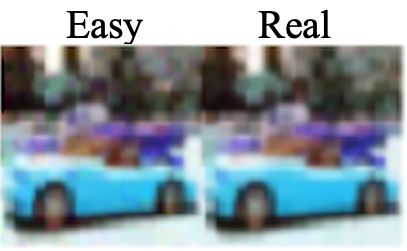}
  \caption{\label{fig:easy} Input class ``0``}
\endminipage
\minipage{0.22\textwidth}
\centering
  \includegraphics[width=0.9\linewidth]{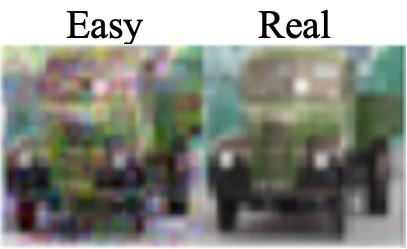}
  \caption{\label{fig:real} Input class ``1``}
\endminipage
\end{figure}
\vspace{-2mm}
As the first step, we trained ResNet-18 one epoch on the CIFAR-10 dataset and collected easy examples for each image by gradient perturbations. Then we took the new randomly initialized ResNet-18 and set up a binary classification task based on the CIFAR-10 dataset. We picked the class ``car`` as ``0`` and class ``truck`` as a class ``1``. Then, each sample was connected with its easy version, see Figure~\ref{fig:easy},~\ref{fig:real}. After training a new model on these ``stacked`` samples, we added noise to one of the image parts and evaluated the accuracy. The idea of this experiment is that: if the easy part of the image is actually easy, the model will learn to classify based only on this part and "do not pay attention" to the other part. As a result, the accuracy drops drastically from $99.0 \rightarrow 0.0$ if we randomize the easy part of the image and do not change $99.0 \rightarrow 99.0$ otherwise. This result fortifies the proposition that neural networks tend to classify based on easy examples rather than the standard ones. 

\textbf{Easy batch normalization:} When we obtain easy examples, we propose to use a separate auxiliary batch normalization for them during the training, similarly as it was done for the adversarial examples in AdvProp. We hypothesize that neural network layers that use the easy batch normalization learn to map hidden state statistics back to parts of the data manifold where the classifier network performs well.

\section{Experiments}
\textbf{Datasets:} We tested our method on FashionMNIST and CIFAR-10 datasets. These datasets consist of the 50000 training samples and 10000 test samples. We used standard data pre-processing for ResNet models without any data augmentations.  

\textbf{Settings:} We employed ResNet-18~\cite{Resnet} and ResNet-50 architectures, used in PyTorch~\cite{NEURIPS2019_9015} AdvProp realization~\cite{AdvPP}. The ResNets were adapted for the CIFAR-10 similarly to~\cite{zhang2019lookahead} with wider channels and more parameters than the original. For adversarial and easy examples generation, we used a 1-step $l_{\inf}$ PGD attack similar to the original AdvProp method. 

For each experiment, training parameters were identical: 1000 training epochs, SGD optimizer, and cosine annealing learning rate scheduler. The learning rate warm-up schedule was equal to 0.1, 0.0005, and 0.0005 values after the 1, 200, and 450 epochs. To test robust accuracy, a 100-step $l_{2}$ PGD attack from robustness library~\cite{robustness} was used. In addition, robustness was evaluated with a standard version of AutoAttack (AA)~\cite{hein} which is a parameter-free and user-independent ensemble of attacks. For AA perturbation size was set to $\varepsilon = 8/255$ for $l_{inf}$ and $\varepsilon = 0.5$ for $l_2$ to keep performance comparable with models presented in RobustBench~\cite{robustbench}.

\textbf{Results:} See results in Table~\ref{tab:table-1},~\ref{tab:table-2}. Easy batch normalization (EBN) meets both goals and improves standard and robust accuracy for most cases, compared to the vanilla AdvProp (AP). The network learns to neutralize adversarial perturbations and maps hidden states back to the data manifold where the network performs well on the training dataset.  
\begin{table}[h!]
\centering
\begin{center}
\begin{scriptsize}
\begin{tabular}{lllllllll}
\hline
Model    & Basic    & $\varepsilon_{l_{2}} 0.1$   & $\varepsilon_{l_{2}} 0.5$  & $\varepsilon_{l_{2}} 1.0$  & AA$_{l_{\inf}}$ & AA$_{l_{2}}$\\ \hline
RN-18 (AP)      & 94.23 & 82.82 & 79.42 & 76.20  & 86.87 & \textbf{90.90}   \\ 
RN-18 (\textbf{EBN})     & \textbf{94.28} & \textbf{86.39} & \textbf{83.27} & \textbf{80.31} & \textbf{86.93} & 90.89 \\
\hline
RN-50 (AP)      & 94.40 & \textbf{62.03} & 60.05 & 57.46   & 85.28 & \textbf{90.27}  \\ 
RN-50 (\textbf{EBN})     & \textbf{94.45} & 63.02 & \textbf{60.65}  & \textbf{58.63} & \textbf{85.38} & 90.20\\ 
 \hline
\end{tabular}
\end{scriptsize}
\end{center}
\vskip -0.1in
\caption{Average results over three random seeds of ResNet (RN) models trained on FashionMNIST}
\label{tab:table-1}
\end{table}
\vskip -0.2in

\begin{table}[h!]
\centering
\begin{center}
\begin{scriptsize}
\begin{tabular}{lllllllll}
\hline
Model      & Basic    & $\varepsilon_{l_{2}} 0.1$   & $\varepsilon_{l_{2}} 0.5$  & $\varepsilon_{l_{2}} 1.0$  & AA$_{l_{\inf}}$ & AA$_{l_{2}}$\\ \hline
RN-18 (AP)      & 93.62    & 36.74          & 29.17          & 22.82                    & 43.12 & \textbf{77.85} \\ 
RN-18 (\textbf{EBN})    & \textbf{93.81}     & \textbf{37.49} & \textbf{30.08} & \textbf{24.39}   & \textbf{43.75} & 77.19 \\
\hline
RN-50 (AP)      & 94.55     & \textbf{38.43}          & 29.39          & 21.96                   & 45.65 & \textbf{78.34}  \\ 
RN-50 (\textbf{EBN})     & \textbf{94.62} & 38.01      & \textbf{30.76}          & \textbf{23.54}    & \textbf{46.47} & 78.05 \\ 
\hline
\end{tabular}
\end{scriptsize}
\end{center}
\vskip -0.1in
\caption{Average results over three random seeds of ResNet (RN) models trained on CIFAR-10}
\label{tab:table-2}
\end{table}
\vskip -0.2in
\section{Conclusion and future works}
We propose a simple remedy to improve the accuracy of classification networks. We experimented with FashionMNIST and CIFAR-10 datasets and showed that our method outperforms basic training settings. Further, we plan to test our method with the recent advancement of the AdvProv method: Fast AdvProp~\cite{mei2022fast} and extend our investigation to more complex datasets with more general robustness measurement~\cite{exval}. In future work we plan to study the latent space representations of the networks trained with easy batch normalization. Especially we plan to evaluate the idea that \textbf{easy examples simplifies decision boundary} and bias neural models to find \textbf{lower rank solutions that generalize well}~\cite{low_rank}.


\newpage

\bibliography{example_paper}
\bibliographystyle{icml2022}

\end{document}